\documentclass[preprint]{article}

\PassOptionsToPackage{numbers, compress}{natbib}

\usepackage{neurips_2026}

\usepackage[utf8]{inputenc} 
\usepackage[T1]{fontenc}    
\usepackage{url}            
\usepackage{booktabs}       
\usepackage{amsfonts}       
\usepackage{nicefrac}       
\usepackage{microtype}      
\usepackage{xcolor}         
\usepackage{graphicx} 
\usepackage{wrapfig}
\usepackage{enumitem}

\usepackage{amsmath}
\usepackage{amssymb}
\usepackage{mathtools}
\usepackage{amsthm}
\usepackage{multirow}
\usepackage{bm}
\usepackage[table]{xcolor}

\definecolor{citeblue}{rgb}{0.21,0.49,0.74}
\usepackage[pagebackref,breaklinks,colorlinks,allcolors=citeblue]{hyperref}

\title{UniSHARP: Universal Sharp Monocular View Synthesis}

\author{%
\vspace{1em}\bfseries
Meixi Song\textsuperscript{1}\quad
Dizhe Zhang\textsuperscript{1}\footnotemark[1]\quad
Hao Ren\textsuperscript{1,2}\quad
Ruiyang Zhang\textsuperscript{1,3}\\
\vspace{1em}\bfseries
Bo Du\textsuperscript{4}\quad
Ming-Hsuan Yang\textsuperscript{5}\quad
Lu Qi\textsuperscript{1,4}\thanks{indicates corresponding author.}\\
\vspace{1em}
\textsuperscript{1} Insta360 Research\quad
\textsuperscript{2} Sun Yat-sen University\quad
\textsuperscript{3} Beihang University\\
\textsuperscript{4} Wuhan University\quad
\textsuperscript{5} University of California, Merced
}

\begin{document}

\maketitle
\begin{figure}[!h]
    \centering
    \includegraphics[width=\linewidth]{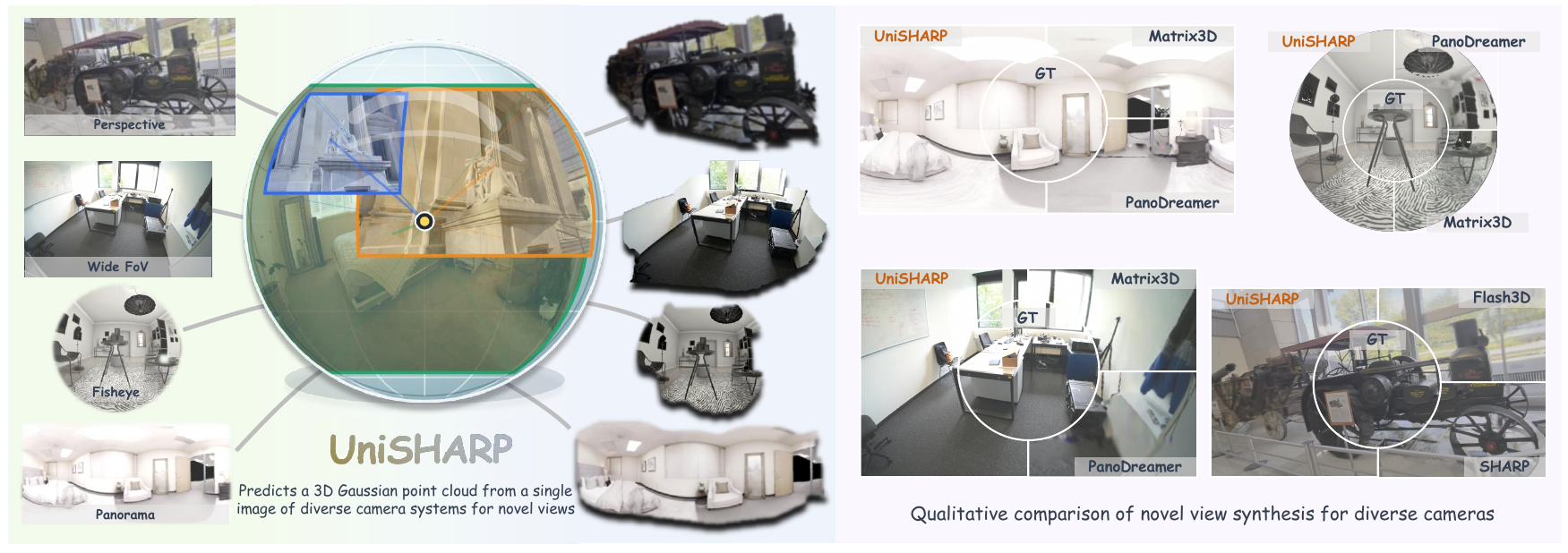}
    \caption{UniSHARP performs monocular novel view synthesis across diverse camera types.
    Given a single image from a perspective, wide-FoV, fisheye, or panoramic camera, UniSHARP predicts a 3D Gaussian point cloud and renders high-quality novel views.
    }
    \label{fig: teaser}
    \vspace{1em}
\end{figure}

\begin{abstract}
In this work, we focus on extending SHARP, the popular photorealistic view synthesis method, for universal monocular rendering across a continuum of camera systems, from conventional perspective cameras to wide-field-of-view, fisheye and omnidirectional panoramic settings.
To overcome the pinhole-specific assumptions of SHARP, our key idea is to align various images in a unified omnidirectional latent space.
Thus, we propose UniSHARP, which performs implicit alignment in both feature and Gaussian spaces.
%
Specifically, Gaussian primitives are arranged along rays and radial distances in a ray-based universal representation, while 2D semantic and 3D spatial features extracted from UniK3D-inspired encoders are jointly decoded to generate the complete Gaussian cloud.
To comprehensively evaluate our method, we construct a benchmark covering diverse imaging systems across various scenes. 
The benchmark is further stratified by field of view (FoV) to enable fine-grained assessment of the universal monocular rendering task.
Extensive experiments on the proposed benchmark demonstrate the effectiveness of UniSHARP, outperforming alternative methods by a large margin.
The project page can be found at: \url{https://insta360-research-team.github.io/Unisharp-website/}

\end{abstract}

\section{Introduction}
Novel view synthesis is a fundamental capability for spatial visual intelligence, enabling captured images to support robotic navigation, AR/VR interaction, immersive telepresence, and 3D content creation~\citep{linhqgs,charatan2024pixelsplat,chen2024mvsplat, song2025d, ren2025prior, wang2026holigs, liu2026mosiv, yue2025roburcdet, ren2026strnet}. 
Albeit the success achieved by neural radiance fields (NeRF)~\citep{mildenhall2021nerf} and 3D Gaussian Splatting (3DGS)~\citep{3dgs}, the monocular view synthesis remains ill-posed due to the severe spatial information loss inherent in a single image.

Recent monocular 3DGS methods, such as SHARP~\citep{mescheder2026sharp} and Flash3D~\citep{szymanowicz2024flash3d}, learn feedforward Gaussian priors from perspective image collections and regress renderable primitives from a single input. 
However, trained primarily on narrow-FoV perspective images, these regressors fail to generalize to diverse imaging systems~\citep{li2024omnigs,chen2025splatter360,zhang2025pansplat,lee2025omnisplat,deng2025selfcaligs}, including wide-FoV, fisheye, and panoramic cameras.
These practical constraints motivate universal monocular novel view synthesis, where a model must infer 3D structure, visibility, and appearance from a single image while remaining applicable to heterogeneous camera models.

%
To address this issue, an intuitive solution is to fine-tune SHARP on wide-FoV or panoramic images. However, since SHARP maps every pixel in normalized space under the pinhole camera assumption, it inherently fails to predict geometry in non-pinhole domain.
Another approach is to re-project images into multiple views, but this introduces additional computational overhead and requires extra processing to handle stitching artifacts.
Thus, one question raised: how can widely used methods such as SHARP be adapted to diverse camera systems in a simple manner?

Inspired by panoramic vision \citep{ge2025airsim360, lin2025depth, zhang2026fly360, feng2025dit360, liu2026omniroam, lin2025one, wang2026panoworld}, we propose UniSHARP, which extends the popular SHARP framework for universal monocular view synthesis via a unified omnidirectional representation across diverse camera systems.
Specifically, UniSHARP performs implicit alignment in the latent rather than the image space along two dimensions.
%
On one hand, a ray-based universal representation organizes Gaussian primitives along rays and radial distances, enabling initialization and refinement in a shared 3D space rather than projection-specific image coordinates.
On the other hand, a unified feature space decodes both 2D semantic embeddings and 3D spatial features, providing complementary appearance and geometry priors for single-image reconstruction.
With such minor modifications, the robustness of UniSHARP to diverse camera systems is enhanced.

For scenarios where camera parameters are unavailable, UniSHARP also supports pose-free monocular inference from a single RGB image.
It uses the predicted ray field to infer the input camera type and recover the rendering geometry, allowing the same feedforward Gaussian predictor to operate without manually provided intrinsics.

To well evaluate our performance, we build a comprehensive benchmark that collects narrow perspective, wide-FoV, fisheye, and panoramic validation data across real-world and simulated scenes. The benchmark combines established validation datasets with OmniRooms, our AirSim-based indoor panoramic dataset, and its projected wide-FoV variant. This FoV-stratified design enables controlled analysis of how rendering quality changes as the camera FoV increases from $60^\circ$ to $360^\circ$.

Our contributions are summarized as follows:
\begin{itemize}[leftmargin=*]
    \item We propose UniSHARP, a universal-camera feedforward 3DGS framework for monocular novel view synthesis across standard perspective, wide-FoV, fisheye, and panoramic inputs. It reformulates SHARP-style Gaussian prediction in ray-distance space and can operate with predicted ray fields when calibrated camera parameters are unavailable.
    \item We develop a feature-space Gaussian prediction pipeline that fuses 2D semantic encodings with 3D spatial features and allocates Gaussians at native input resolution, preserving geometric priors and high-frequency image details without camera-specific resizing.
    \item We introduce panoramic-specific adaptations, including spherical Gaussian initialization and distortion-aware probabilistic dropout, to regularize Gaussian distributions under severe equirectangular projection distortion.
    %
    %
    \item We introduce a FoV-stratified benchmark spanning perspective, wide-FoV, fisheye, and panoramic cameras. We validate UniSHARP on this benchmark, demonstrating state-of-the-art rendering quality and strong cross-camera generalization. 
\end{itemize}

\begin{figure}[t]
    \centering
    \includegraphics[width=1\linewidth]{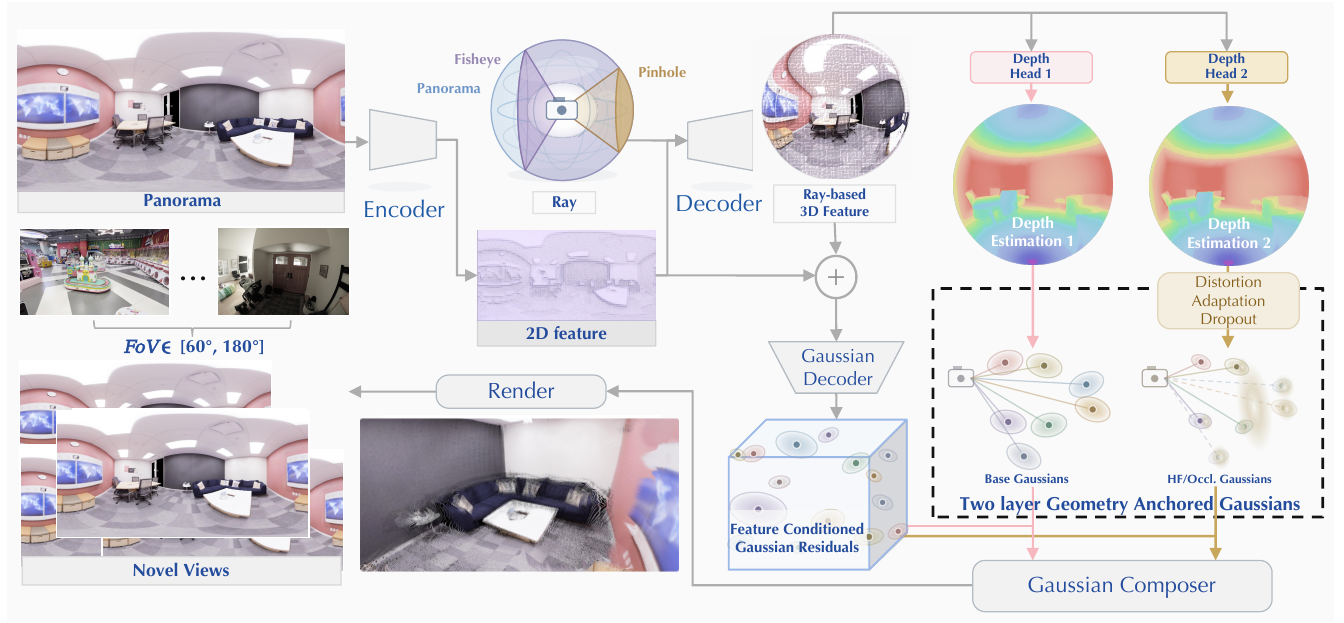}
    \caption{UniSHARP pipeline for universal-camera monocular novel view synthesis.
    Given a single source image, UniSHARP estimates ray-distance geometry and multi-scale features, initializes two-layer Gaussians in ray-distance space, predicts Feature Conditioned Gaussian residuals, and renders target views with the unified Gaussian representation across perspective, wide-FoV, fisheye, and panoramic cameras.
    }
    \label{figpipeline}
\end{figure}

\section{Related work}

\noindent\textbf{Multi-image novel view synthesis.} Multi-image NVS reconstructs scenes from several posed or nearby observations by exploiting cross-view consistency.
Neural radiance fields established continuous scene representations for photorealistic rendering~\citep{mildenhall2021nerf}, while later anti-aliased variants improved unbounded and high-resolution reconstruction~\citep{barron2022mipnerf360,barron2023zipnerf}. 
3D Gaussian Splatting replaced expensive volume rendering with explicit primitives and real-time rasterization~\citep{3dgs}.
Feedforward methods further reduce per-scene optimization through learned image-based rendering and multi-view stereo cost volumes~\citep{wang2021ibrnet,chen2021mvsnerf}, while recent sparse-view Gaussian models predict 3DGS representations from image pairs or posed image sets~\citep{charatan2024pixelsplat,chen2024mvsplat,xu2025depthsplat}. 
Large reconstruction models and pose-free systems extend this trend to larger baselines or unconstrained captures~\citep{jiang2025anysplat,hong2025pf3plat,chen2024lara,zhang2024gslrm}. 
Despite their strong quality, these methods fundamentally depend on camera poses, correspondences, or multi-view feature aggregation. 
UniSHARP instead targets the strictly monocular setting and predicts a renderable Gaussian representation from a single image.

\noindent\textbf{Single-image novel view synthesis.} Single-image NVS must infer geometry, appearance, and visibility from priors rather than direct triangulation. 
Early learning-based approaches predicted neural radiance fields or multiplane images from a single input~\citep{yu2021pixelnerf,tucker2020singleview}, and other methods used layered geometry, inpainting, or adaptive MPI layouts to handle disocclusion~\citep{wiles2020synsin,jampani2021slide,han2022adampi,khan2023tmpi}. 
Larger reconstruction and synthesis models show that strong feedforward networks can infer plausible 3D structure from a single photograph~\citep{hong2023lrm,jin2025lvsm}. 
More recently, monocular 3DGS methods directly regress explicit Gaussian primitives, enabling efficient rendering from single images~\citep{mescheder2026sharp,szymanowicz2024flash3d,szymanowicz2024splatter}. 
Generative methods improve extrapolation to larger camera motion through diffusion priors~\citep{sargent2024zeronvs,liang2025wonderland,ren2025gen3c}, but they often trade rendering speed or geometric explicitness for generative flexibility. 
These works establish single-image NVS as a compelling direction, yet they primarily assume perspective imagery. 
UniSHARP extends monocular 3DGS to a unified camera setting covering perspective, wide-FoV, fisheye, and panoramic inputs.

\noindent\textbf{Wide-FoV novel view synthesis.} Wide-FoV NVS introduces projection distortion, nonuniform angular sampling, and nontrivial boundary topology that are absent in perspective images. 
Universal monocular geometry estimators show that ray-based representations are important for handling arbitrary cameras~\citep{piccinelli2025unik3d}. 
In reconstruction, recent 3DGS systems adapt the rasterizer or camera model to omnidirectional and fisheye inputs~\citep{li2024omnigs,liao2024fisheyegs}, and self-calibrating variants jointly optimize camera models, poses, and Gaussian scenes for wide-FoV captures~\citep{deng2025selfcaligs,huang2025scomnigs,yang2026directfisheyegs}. 
In parallel, panoramic feedforward methods use spherical radiance fields, spherical cost volumes, Gaussian pyramids, or Yin-Yang grids for 360-degree synthesis~\citep{chen2025splatter360,zhang2025pansplat,lee2025omnisplat,chen2023panogrf}. 
These methods address important non-perspective settings, but they commonly rely on optimized scenes, calibrated captures, or multiple panoramic observations. 
UniSHARP instead unifies perspective, wide-FoV, fisheye, and panoramic monocular NVS in a single feedforward 3DGS model.

\section{Method}
Given a single source image $I_s \in \mathbb{R}^{3 \times H \times W}$, UniSHARP predicts a set of 3D Gaussian primitives to enable high-fidelity novel-view synthesis.
Unlike conventional feedforward methods that regress Gaussian attributes in the image plane, UniSHARP decouples camera projection from scene representation via a unified ray-distance space. This is achieved by: (1) introducing a ray-based universal representation for heterogeneous cameras (Sec.~\ref{method:ray}); (2) composing the scene using Geometry Anchored Gaussians and Feature Conditioned Gaussian residuals (Sec.~\ref{method:model}); (3) employing a mixed-camera training strategy to supervise rendering across diverse projection types (Sec.~\ref{method:training}); and (4) enabling pose-free inference by estimating camera geometry from predicted rays (Sec.~\ref{method:posefree}).

\subsection{Ray Based Universal Representation}
\label{method:ray}

Standard image-plane coordinates fail to generalize across heterogeneous projections because a single-pixel displacement corresponds to vastly different angular changes in perspective, fisheye, or panoramic views. To address this, UniSHARP adopts a unified ray-distance space inspired by UniK3D~\citep{piccinelli2025unik3d}. By decoupling viewing direction from scene range, we ensure that Gaussian primitives defined by 3D centers, covariances, and appearance are optimized in a consistent metric space instead of being tied to projection-specific image grids.

Formally, let $\Omega=\{1,\ldots,H\}\times\{1,\ldots,W\}$ denote the pixel domain. UniSHARP uses a predicted per-pixel unit ray field
\begin{equation}
\mathbf{r}_p
\in
\mathbb{S}^2,
\qquad
\left\|\mathbf{r}_p\right\|_2=1,
\label{eq:ray_field}
\end{equation}
and a radial distance $d_p>0$ from the camera center. The corresponding 3D point is then \(\mathbf{x}_p = d_p \mathbf{r}_p\).

This formulation provides a universal coordinate system where Gaussian attributes (placement, scale, and color) are defined consistently across diverse camera models. By measuring spatial footprints along rays rather than in the rasterized plane, UniSHARP enables robust initialization and refinement of Gaussian scenes regardless of the input projection type.
\subsection{Model Design}
\label{method:model}

Building upon the unified ray system, UniSHARP first constructs Geometry Anchored Gaussians  (\ref{method:geometry}), providing a camera-unified gaussian space initialization.
It then predicts Feature Conditioned Gaussian residuals (\ref{method:feature}) from 2D semantic and 3D ray-based features, composing them with the anchors to obtain the final renderable Gaussians. The overall pipeline is illustrated in Figure~\ref{figpipeline}.

\subsubsection{Geometry Anchored Gaussians}
\label{method:geometry}

For each input image, we construct two-layer Geometry Anchored Gaussians on a native ray grid.
Let $H_g$ and $W_g$ be the Gaussian grid resolution.
At pixel $p$ and layer $\ell$, each Geometry Anchored Gaussian is represented as
\begin{equation}
\mathcal{B}_{p,\ell}
=
\left\{
\mathbf{r}_{p}, \rho_{p,\ell},
\mathbf{s}_{p,\ell}^{0},
\mathbf{q}^{0},
\mathbf{c}_{p}^{0},
\alpha_{\ell}^{0}
\right\},
\label{eq:base_gaussian}
\end{equation}
where $\mathbf{r}_{p}$ is the unit ray, $\rho_{p,\ell}$ is inverse radial distance, $\mathbf{s}_{p,\ell}^{0}$ is the base scale, $\mathbf{q}^{0}$ is the identity quaternion, $\mathbf{c}_{p}^{0}$ is obtained from the input color, and $\alpha_{\ell}^{0}$ is opacity initialization.
The first layer aligns with the visible surface, while the second layer captures disocclusions and high-frequency structures beyond a single surface hypothesis.
This second radial layer is predicted by an additional depth head same as the UniK3D radial head, giving the geometry-anchored representation a separate distance hypothesis that can specialize through rendering supervision and regularization.

By allocating primitives according to native ray sampling, UniSHARP enables resolution-adaptive construction that preserves angular consistency and high-frequency details across wide-FoV and panoramic inputs without the distortions of a fixed grid.

\subsubsection{Feature Conditioned Gaussian Residuals}
\label{method:feature}

While Geometry Anchored Gaussians provide a camera-unified layout, UniSHARP further predicts Feature Conditioned Gaussian residuals to incorporate the semantic context and geometric priors necessary for high-fidelity synthesis.
Unlike conventional monocular predictors that feed RGB images and depth images directly into a decoder, UniSHARP predicts residuals within a unified space by fusing 2D semantic image features with 3D ray-based geometric features.

For each geometry-anchored location, a Gaussian decoder predicts a residual tensor $\Delta\in\mathbb{R}^{B \times 14 \times L \times H_g \times W_g}$, whose channels correspond to tangent-plane center offsets, inverse-distance, scale, quaternion, color, and opacity residuals.
The final Gaussian is obtained by composing the anchor and residual:
\begin{equation}
\mathcal{G}_{p,\ell}
=
\operatorname{Compose}(\mathcal{B}_{p,\ell},\Delta_{p,\ell})
=
\{\boldsymbol{\mu}_{p,\ell},\mathbf{s}_{p,\ell},\mathbf{q}_{p,\ell},\mathbf{c}_{p,\ell},\alpha_{p,\ell}\}.
\label{eq:gaussian_compose}
\end{equation}

\subsection{Training Strategy and Objective}
\label{method:training}

\noindent\textbf{Mixed-camera training.}
UniSHARP is trained under a mixed-camera regime, jointly optimizing perspective, wide-FoV, fisheye, and panoramic data within a single model. 
During training, a weighted sampler draws source-target image pairs from all supported datasets and groups each mini-batch by dataset for efficient collation and rendering. 
Although these data differ in image formation process, FoV, and valid target regions, UniSHARP does not introduce camera-specific branches. 
Instead, each sample is converted into the same ray-based training interface, and all camera types share a unified network architecture. 
As a result, UniSHARP learns a camera-unified Gaussian representation that transfers supervision across heterogeneous cameras while avoiding separate predictors for individual camera models.

\noindent\textbf{Panoramic distortion adaptation.}
Equirectangular panoramas oversample polar regions because pixels near the poles correspond to narrower solid angles than those at the equator. To address this, we apply a latitude-dependent probabilistic dropout on the second Gaussian layer during training:
\begin{equation}
p_y
=
1-
\frac{
\max(\cos\theta_y,0)
}{
\max_{y'} \max(\cos\theta_{y'},0)
},
\label{eq:pano_dropout_prob}
\end{equation}
where $\theta_y$ is the latitude of row $y$.
The first layer is always preserved to maintain visible surface coverage.
While the first layer is preserved for surface coverage, the second is then selectively suppressed via a Bernoulli mask $m_{p,2} \sim \text{Bernoulli}(p_y)$ that biases opacity residuals. This approach shifts panoramic distortion adaptation from a specialized prediction branch to a training-time allocation strategy.

\noindent\textbf{Objective.}
Let $s$ and $t$ denote source and target views.
The training objective includes appearance supervision, depth supervision, and Gaussian regularization.
We use $\hat{\mathbf{I}}_v$, $\hat{\mathbf{A}}_v$, and $\hat{\mathbf{D}}_v$ for rendered color, opacity, and distance at view $v$, and $\mathbf{I}_v$ and $\mathbf{D}_v$ for image and depth supervision.
Appearance supervision encourages the accumulated opacity to cover valid pixels, and applies a perceptual term on target views where novel view artifacts are most visible.
\begin{equation}
\mathcal{L}_{\mathrm{app}} =
\lambda_c
\sum_{v\in\{s,t\}}
\left\|\hat{\mathbf{I}}_v-\mathbf{I}_v\right\|_1
+
\lambda_a
\sum_{v\in\{s,t\}}
\mathrm{BCE}(\hat{\mathbf{A}}_v,\mathbf{1})
+
\lambda_p
\Phi(\hat{\mathbf{I}}_t,\mathbf{I}_t),
\label{eq:appearance_loss}
\end{equation}
where $\Phi$ is a perceptual loss computed from deep features and Gram statistics.

Depth supervision anchors both the source geometry used to initialize Gaussians and the target geometry produced after splatting.
\begin{equation}
\mathcal{L}_{\mathrm{dep}}
=
\lambda_d
\left(
\left\|
\tilde{\mathbf{D}}_s^{-1}
-
\mathbf{D}_s^{-1}
\right\|_1
+
\left\|
\hat{\mathbf{D}}_t^{-1}
-
\mathbf{D}_t^{-1}
\right\|_1
\right),
\label{eq:depth_loss}
\end{equation}

where $\tilde{\mathbf{D}}_s$ is the first source radial layer and $\hat{\mathbf{D}}_t$ is the rendered target distance.

Gaussian regularization stabilizes degrees of freedom that are weakly constrained by photometric loss.
It includes total variation on the second radial layer, floater suppression near abrupt first-layer inverse-distance changes, and multi-scale Sobel alignment in log-distance space:
\begin{align}
\mathcal{L}_{\mathrm{geo}}
&=
\lambda_{\mathrm{tv}}
\left\langle
\left\|
\nabla \hat{\mathbf{D}}_{s,2}^{-1}
\right\|_1
\right\rangle
+
\lambda_g
\left\langle
\hat{\mathbf{O}}
\left(
1-\exp
\left(
-\left[
\left\|\nabla \hat{\mathbf{D}}_{s,1}^{-1}\right\|_1-\tau
\right]_+/\sigma
\right)
\right)
\right\rangle
\notag\\
&\quad
+
\lambda_{\mathrm{gi}}
\frac{1}{K}
\sum_{v\in\{s,t\}}
\sum_{k=1}^{K}
\left\langle
\left\|
\nabla_{\mathrm{Sobel}}
\mathcal{P}_k
\left(
\log \hat{\mathbf{D}}_v-\log \mathbf{D}_v
\right)
\right\|_2
\right\rangle,
\label{eq:geometry_regularization}
\end{align}
where $\hat{\mathbf{O}}$ is the predicted Gaussian opacity, $K$ is the number of depth pyramid scales, $\mathcal{P}_k$ downsamples its input to the $k$-th scale, and $[x]_+=\max(x,0)$.
The three terms respectively smooth the inverse distance of the second source layer, suppress opaque floaters around sharp first-layer distance discontinuities, and align rendered and supervised distance edges with multi-scale Sobel gradients.

For equirectangular panoramas, horizontal finite differences use circular boundary handling to respect the wrap-around topology.
The full training objective is
\begin{equation}
\begin{aligned}
\mathcal{L}
=
\mathcal{L}_{\mathrm{app}}
+
\mathcal{L}_{\mathrm{dep}}
+
\mathcal{L}_{\mathrm{geo}}.
\end{aligned}
\label{eq:full_objective}
\end{equation}

\subsection{Extension to Pose-Free Model}
\label{method:posefree}

UniSHARP naturally supports a calibrated setting where the input camera model and intrinsics are known.
For in-the-wild deployment, however, a user may provide only a single RGB image.
We therefore introduce a pose-free model that replaces external calibration with camera geometry recovered from the predicted UniK3D ray field.
Specifically, we determine the camera model from the angular coverage of the predicted rays and then recover the corresponding rendering geometry.
For perspective and fisheye inputs, the ray field is converted into a compact parametric camera by fitting pinhole intrinsics or Fisheye parameters, while panoramic inputs use the deterministic spherical camera model.
The recovered camera is used consistently for ray-distance Gaussian initialization and novel-view rendering, so the Gaussian predictor remains shared with the calibrated model rather than becoming a separate branch.
This design lets UniSHARP render orbit and forward-motion views from an uncalibrated image while preserving the same feedforward inference pipeline.

\begin{table}[t]
\centering
\caption{Composition of the proposed FoV stratified benchmark for universal-camera monocular novel view synthesis. Validation pairs are grouped by effective FoV and projection type, and sample counts denote evaluated source-target pairs.}
\vspace{-0.5em}
\label{tab:benchmark}
\resizebox{\linewidth}{!}{
\begin{tabular}{lllc}
\toprule
Camera group & FoV range & Validation datasets & Validation samples \\
\midrule
Perspective & $60^\circ$--$90^\circ$ & \begin{tabular}[c]{@{}l@{}}DL3DV~\citep{ling2024dl3dv}, RealEstate10K~\citep{zhou2018stereo},\\ Tanks and Temples~\citep{knapitsch2017tanks}, WildRGB-D~\citep{xia2024wildrgbd}\end{tabular} & 36,873 \\
Wide FoV & $90^\circ$--$140^\circ$ & OmniRooms-Wide (ours), projected from OmniRooms & 10,692 \\
Fisheye & $140^\circ$--$180^\circ$ & ScanNet++ Fisheye~\citep{yeshwanth2023scannetpp} & 14,163 \\
Panorama & $360^\circ$ & Replica~\citep{straub2019replica}, HM3D~\citep{ramakrishnan2021hm3d}, OmniRooms (ours) & 42,754 \\
\bottomrule
\end{tabular}}
\end{table}

\section{Experiments}
\label{exp:setting}

\noindent\textbf{Training.}
We train one unified model on a mixture of perspective, wide-FoV, fisheye, and panoramic datasets.
For perspective images, we use RealEstate10K~\citep{zhou2018stereo}, DL3DV~\citep{ling2024dl3dv}, and WildRGB-D~\citep{xia2024wildrgbd}.
For wide-FoV images, we use OmniRooms-Wide, which is projected from our simulated indoor panoramic dataset.
For fisheye images, we use ScanNet++ Fisheye~\citep{yeshwanth2023scannetpp}.
For panoramic images, we use HM3D panorama datasets~\citep{ramakrishnan2021hm3d} and OmniRooms, our simulated indoor ERP dataset constructed with the AirSim platform~\citep{simulator} by rendering collision-free camera trajectories in synthetic indoor scenes.
All datasets are converted into source-target training pairs under their native camera models.
At each iteration, we first sample a dataset according to a fixed dataset-level distribution and then draw a batch from that dataset.
For datasets without ground-truth depth, we use UniK3D~\citep{piccinelli2025unik3d} to generate pseudo depth labels.
Additional implementation details are provided in Appendix~\ref{app:implementation_details}.

\begin{figure}[t]
    \centering
    \includegraphics[width=1\linewidth]{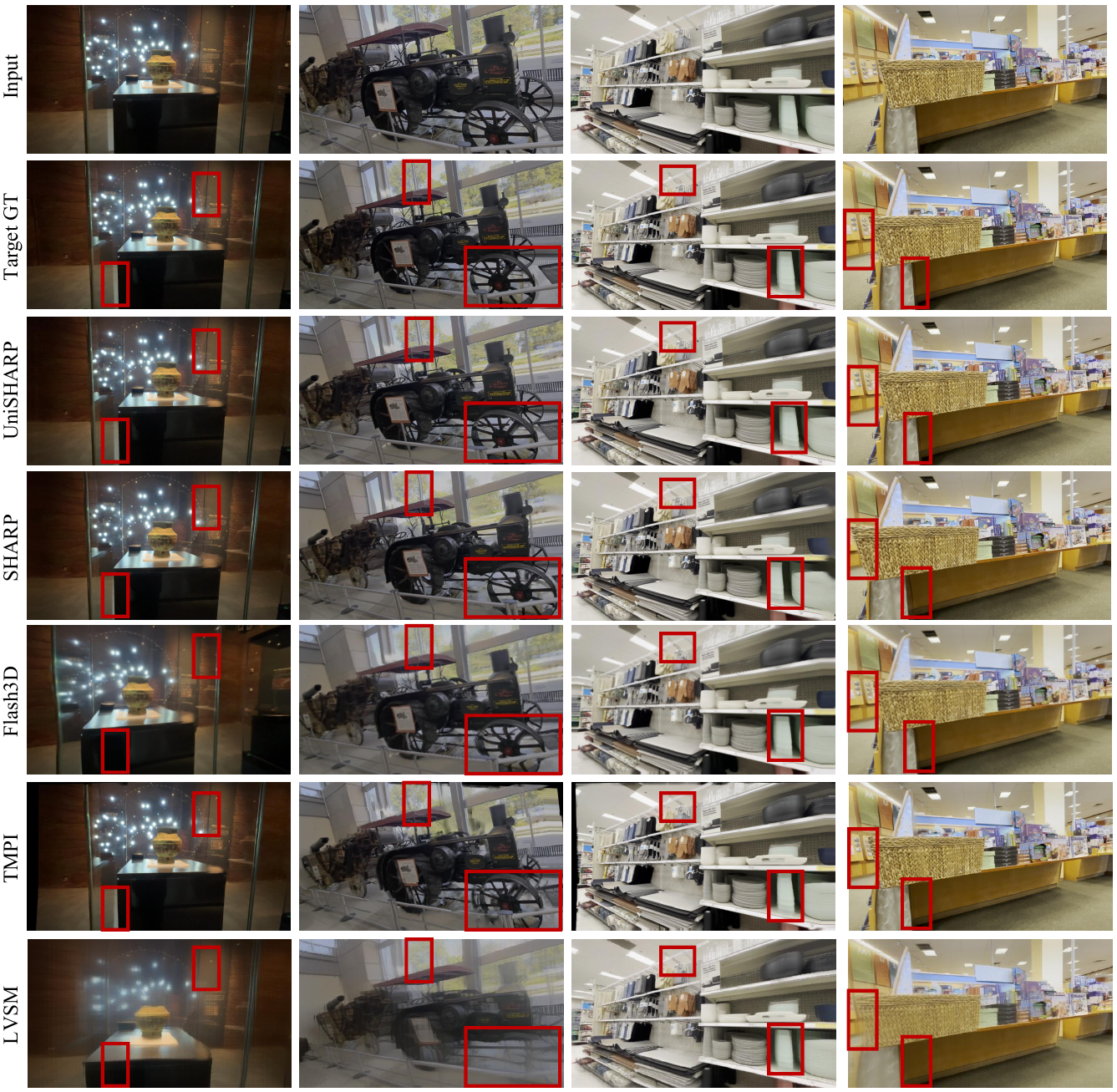}
    \caption{Qualitative comparison on perspective novel view synthesis.
    Given a single source image, UniSHARP produces sharper target-view geometry and fewer disocclusion artifacts than perspective monocular baselines, while preserving view-consistent scene structure.
    }
    \label{pin}
    \vspace{-1em}
\end{figure}

\begin{table*}[t]
\centering
\caption{Quantitative comparison on perspective datasets included in the mixed training distribution. The best results are shown in red and the second-best results are in orange.}
\label{tab:perspective_indomain}
\resizebox{\linewidth}{!}{
\begin{tabular}{lccccccccc}
\toprule
\multirow{2}{*}{Method} & \multicolumn{3}{c}{WildRGB-D~\citep{xia2024wildrgbd}} & \multicolumn{3}{c}{DL3DV~\citep{ling2024dl3dv}} & \multicolumn{3}{c}{RealEstate10K~\citep{zhou2018stereo}} \\
\cmidrule(lr){2-4}\cmidrule(lr){5-7}\cmidrule(lr){8-10}
& PSNR$\uparrow$ & SSIM$\uparrow$ & LPIPS$\downarrow$ & PSNR$\uparrow$ & SSIM$\uparrow$ & LPIPS$\downarrow$ & PSNR$\uparrow$ & SSIM$\uparrow$ & LPIPS$\downarrow$ \\
\midrule
TMPI~\citep{khan2023tmpi} & 15.969 & 0.493 & 0.330 & 16.002 & 0.490 & 0.340 & 21.707 & 0.768 & 0.142 \\
LVSM~\citep{jin2025lvsm} & 15.970 & 0.362 & 0.279 & 15.848 & 0.446 & 0.374 & 22.109 & 0.731 & 0.138 \\
Flash3D~\citep{szymanowicz2024flash3d} & 17.378 & 0.553 & 0.395 & \cellcolor{orange!25}18.418 & \cellcolor{orange!25}0.574 & \cellcolor{red!25}0.179 & 23.573 & 0.785 & 0.172 \\
SHARP~\citep{mescheder2026sharp} & \cellcolor{orange!25}18.300 & \cellcolor{orange!25}0.568 & \cellcolor{orange!25}0.212 & 17.368 & 0.545 & 0.286 & \cellcolor{orange!25}24.215 & \cellcolor{orange!25}0.789 & \cellcolor{orange!25}0.101 \\
\midrule
UniSHARP & \cellcolor{red!25}\textbf{21.556} & \cellcolor{red!25}\textbf{0.674} & \cellcolor{red!25}\textbf{0.143} & \cellcolor{red!25}\textbf{19.468} & \cellcolor{red!25}\textbf{0.594} & \cellcolor{orange!25}\textbf{0.196} & \cellcolor{red!25}\textbf{24.495} & \cellcolor{red!25}\textbf{0.795} & \cellcolor{red!25}\textbf{0.087} \\
\bottomrule
\end{tabular}}
\vspace{-1em}
\end{table*}

\subsection{Benchmark \& Metrics.}
\label{sec:benchmark}
We evaluate UniSHARP on monocular NVS across perspective, wide-FoV, fisheye, and panoramic cameras. To address the limitations of evaluations tied to single projection families, we introduce a FoV stratified benchmark (Table~\ref{tab:benchmark}). This benchmark provides a unified protocol to diagnose model behavior as camera geometry scales from narrow perspective to full $360^\circ$ equirectangular inputs.

\noindent\textbf{OmniRooms.}
OmniRooms is a simulated indoor ERP dataset collected via AirSim~\citep{simulator}, with OmniRooms-Wide derived by projecting these panoramas into $130^\circ$ equidistant fisheye views. For each anchor point on a $0.5$m voxel grid, we render one central camera and 29 others randomly sampled within a local axis-aligned cube of edge length $30$~cm around the source camera. Each frame is rendered as a $1024\times2048$ ERP image and all cameras share a fixed orientation.

\begin{figure}[t]
    \centering
    \includegraphics[width=1\linewidth]{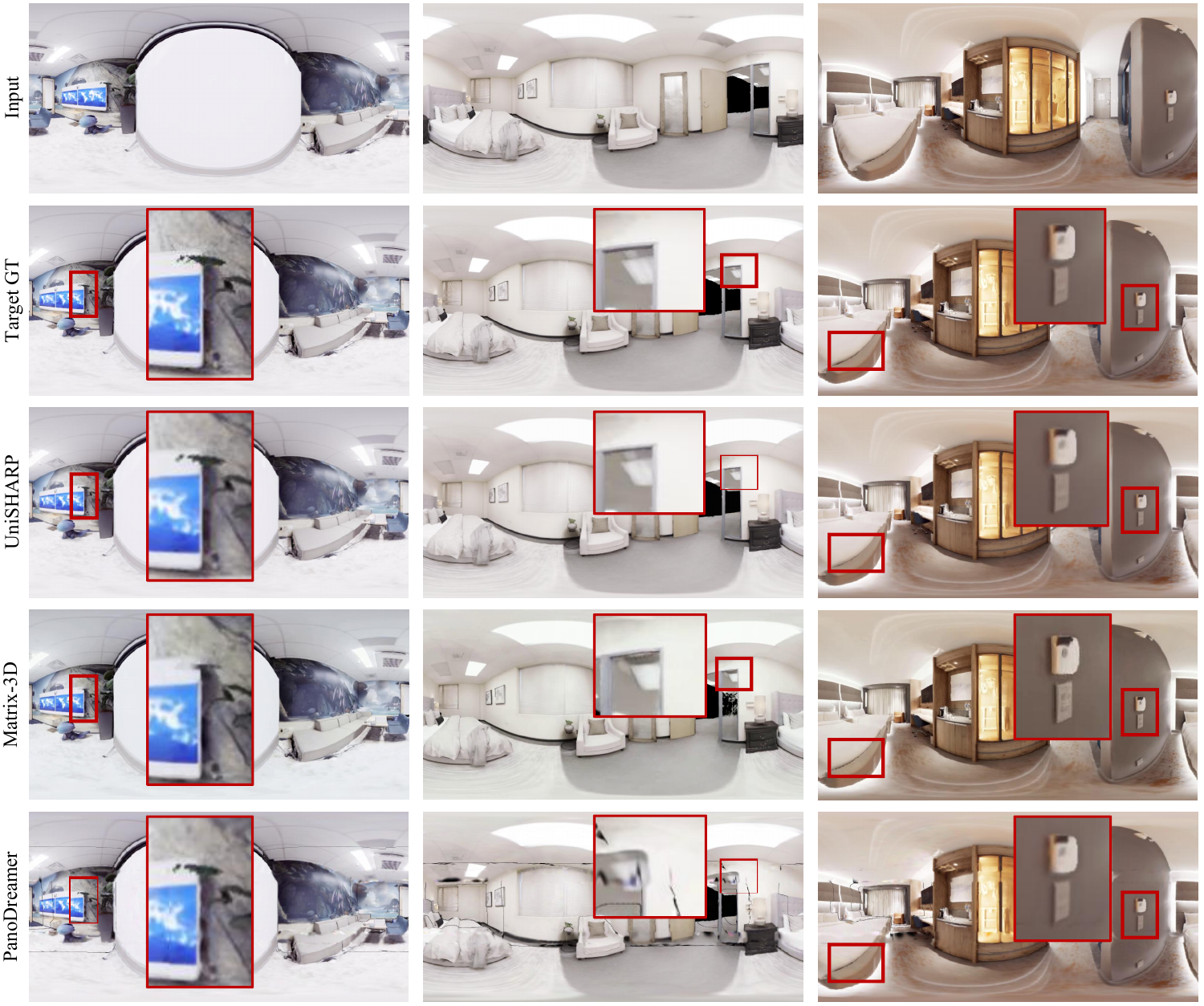}
    \caption{Qualitative comparison on panorama novel view synthesis.
    UniSHARP reconstructs coherent Gaussian geometry from a single panoramic input and renders sharper target views with fewer distortion-induced artifacts.
    }
    \label{fig:pano_qual}
    \vspace{-0.5em}
\end{figure}

\begin{table*}[t]
\centering
\caption{Quantitative comparison on panoramic novel view synthesis. Results are reported on real and simulated ERP datasets using PSNR$\uparrow$, SSIM$\uparrow$, and LPIPS$\downarrow$; best results are shown in red.}
\label{tab:panorama_results}
\resizebox{\linewidth}{!}{
\begin{tabular}{lccccccccc}
\toprule
& \multicolumn{3}{c}{HM3D~\citep{ramakrishnan2021hm3d}} & \multicolumn{3}{c}{OmniRooms} & \multicolumn{3}{c}{Replica~\citep{straub2019replica}} \\
\cmidrule(lr){2-4}\cmidrule(lr){5-7}\cmidrule(lr){8-10}
Method & PSNR$\uparrow$ & SSIM$\uparrow$ & LPIPS$\downarrow$ & PSNR$\uparrow$ & SSIM$\uparrow$ & LPIPS$\downarrow$ & PSNR$\uparrow$ & SSIM$\uparrow$ & LPIPS$\downarrow$ \\
\midrule
PanoDreamer~\citep{paliwal2025panodreamer} & 21.856 & 0.708 & 0.152 & 18.567 & 0.692 & 0.240 & 21.289 & 0.793 & 0.137 \\
Matrix3D~\citep{lu2025matrix3d} & \cellcolor{orange!25} 23.398 & \cellcolor{orange!25} 0.793 & \cellcolor{orange!25} 0.114 & \cellcolor{orange!25} 21.635 & \cellcolor{orange!25} 0.799 & \cellcolor{orange!25} 0.144 & \cellcolor{orange!25} 25.379 & \cellcolor{orange!25} 0.887 & \cellcolor{orange!25} 0.069 \\
\midrule
UniSHARP & \cellcolor{red!25}\textbf{29.244} & \cellcolor{red!25}\textbf{0.895} & \cellcolor{red!25}\textbf{0.065} & \cellcolor{red!25}\textbf{24.004} & \cellcolor{red!25}\textbf{0.856} & \cellcolor{red!25}\textbf{0.123} & \cellcolor{red!25}\textbf{30.182} & \cellcolor{red!25}\textbf{0.933} & \cellcolor{red!25}\textbf{0.038} \\
\bottomrule
\end{tabular}}
\vspace{-1em}
\end{table*}

\noindent\textbf{Benchmark details.}
To align with single-source monocular NVS, we restrict target views to locally reachable positions: we require $>60\%$ source-target overlap, a camera-center distance $<0.5$m, and an image-index gap $<10$. This design focuses evaluation on geometry inference under meaningful motion rather than unconstrained long-range hallucination. The protocol serves as a unified testbed for universal-camera NVS, enabling a direct diagnostic of how rendering quality scales across perspective, wide-FoV, fisheye, and 360-degree projections. We evaluate in a single-source, multi-target setting, using the first frame of each sequence as the source and the subsequent ten frames as targets. Results are reported using PSNR, SSIM, and LPIPS, averaged over all valid target views.

\noindent\textbf{Baselines.}
For perspective novel view synthesis, we compare with representative single-image 3D Gaussian regressors, including SHARP~\citep{mescheder2026sharp} and Flash3D~\citep{szymanowicz2024flash3d}, as well as methods based on different scene representations, including LVSM~\citep{jin2025lvsm}, a large view synthesis model with minimal 3D inductive bias, and TMPI~\citep{khan2023tmpi}, a tiled multiplane-image representation for practical 3D photography.
For wide-FoV, fisheye, and panoramic novel view synthesis, we compare with methods that cover different generation paradigms: PanoDreamer~\citep{paliwal2025panodreamer}, an optimization-based single-image-to-360 scene method using diffusion and 3DGS, and Matrix3D~\citep{lu2025matrix3d}, a diffusion-based video generation model.

\begin{table*}[t]
\centering
\caption{Quantitative comparison on wide-FoV and fisheye novel view synthesis. Results are reported on OmniRooms-Wide and ScanNet++ Fisheye.}
\label{tab:fisheye_results}
\fontsize{9}{10}\selectfont
\setlength{\tabcolsep}{0pt}
\renewcommand\arraystretch{0.9}
\begin{tabular*}{\linewidth}{@{\extracolsep{\fill}}lcccccc}
\toprule
& \multicolumn{3}{c}{ScanNet++ Fisheye~\citep{yeshwanth2023scannetpp}} & \multicolumn{3}{c}{OmniRooms-Wide} \\
\cmidrule(lr){2-4}\cmidrule(lr){5-7}
Method & PSNR$\uparrow$ & SSIM$\uparrow$ & LPIPS$\downarrow$ & PSNR$\uparrow$ & SSIM$\uparrow$ & LPIPS$\downarrow$ \\
\midrule
PanoDreamer~\citep{paliwal2025panodreamer} & 15.131 & 0.682 & 0.383 & 18.062 & 0.760 & 0.199 \\
Matrix3D~\citep{lu2025matrix3d} & \cellcolor{orange!25}16.382 & \cellcolor{orange!25}0.690 & \cellcolor{orange!25}0.371 & \cellcolor{orange!25} 18.881 & \cellcolor{orange!25} 0.795 & \cellcolor{orange!25} 0.173 \\
\midrule
UniSHARP & \cellcolor{red!25}\textbf{20.660} & \cellcolor{red!25}\textbf{0.771} & \cellcolor{red!25}\textbf{0.184} & \cellcolor{red!25}\textbf{25.243} & \cellcolor{red!25}\textbf{0.854} & \cellcolor{red!25}\textbf{0.076} \\
\bottomrule
\end{tabular*}
\vspace{-2em}
\end{table*}

\subsection{Qualitative and Quantitative Evaluation}

\noindent\textbf{Perspective performance.}

\begin{wraptable}[9]{r}{0.55\columnwidth}
    \vspace{-1.7em}
    \centering
    \caption{Zero-shot perspective evaluation on Tanks and Temples~\citep{knapitsch2017tanks}.}
    \fontsize{9}{10}\selectfont
    \setlength{\tabcolsep}{9pt}
    \renewcommand\arraystretch{0.9}
    \begin{tabular}{lccc}
        \toprule
        Method & PSNR$\uparrow$ & SSIM$\uparrow$ & LPIPS$\downarrow$ \\
        \midrule
        TMPI~\citep{khan2023tmpi} & 13.623 & 0.399 & 0.440 \\
        LVSM~\citep{jin2025lvsm} & 14.575 & 0.379 & 0.476 \\
        Flash3D~\citep{szymanowicz2024flash3d} & \cellcolor{orange!25}15.985 & \cellcolor{red!25}\textbf{0.511} & 0.312 \\
        SHARP~\citep{mescheder2026sharp} & 15.964 & \cellcolor{orange!25}0.502 & \cellcolor{orange!25}0.301 \\
        UniSHARP & \cellcolor{red!25}\textbf{16.315} & 0.498 & \cellcolor{red!25}\textbf{0.282} \\
        \bottomrule
    \end{tabular}
    \label{tab:perspective_outdomain}
    \vspace{-1em}
\end{wraptable}
Table~\ref{tab:perspective_indomain} reports the in-domain results on RealEstate10K, DL3DV, and WildRGB-D.
UniSHARP achieves the best PSNR and SSIM across the perspective datasets and the best or second-best LPIPS, showing that universal-camera training preserves strong standard perspective rendering.
The gains are consistent across real-estate, object-centric, and in-the-wild scenes, indicating that the shared ray-distance Gaussian representation improves fidelity without overfitting to a single perspective dataset.

Table~\ref{tab:perspective_outdomain} evaluates out-of-domain generalization on the Tanks and Temples dataset.
UniSHARP obtains the best PSNR and LPIPS compared with the strongest baselines.
The SSIM remains close to Flash3D, suggesting that the unified representation preserves cross-dataset generalization while improving overall reconstruction fidelity.

\noindent\textbf{Wide-FoV, fisheye, and panoramic performance.}

\begin{wraptable}[8]{r}{0.55\columnwidth}
\vspace{-1.7em}
\centering
\caption{Pose-free evaluation on WildRGB-D. Ours uses the available camera parameters, while Ours (pose-free) estimates camera geometry from predicted rays.}
\label{tab:posefree_wildrgbd}
\fontsize{9}{10}\selectfont
\setlength{\tabcolsep}{8pt}
\renewcommand\arraystretch{0.95}
\begin{tabular}{lccc}
\toprule
Setting & PSNR$\uparrow$ & SSIM$\uparrow$ & LPIPS$\downarrow$ \\
\midrule
Ours (pose-free) & 20.850 & 0.647 & 0.157 \\
Ours & \textbf{21.556} & \textbf{0.674} & \textbf{0.143} \\
\bottomrule
\end{tabular}
\end{wraptable}
Tables~\ref{tab:panorama_results} and Table~\ref{tab:fisheye_results} evaluate UniSHARP on non-perspective cameras.
On OmniRooms-Wide, UniSHARP consistently improves PSNR, SSIM, and LPIPS over both baselines.
This shows that the ray-distance Gaussian representation remains effective for wide-FoV inputs, where the model must handle stronger projection distortion and larger angular coverage than standard perspective images.
On ScanNet++ Fisheye, UniSHARP also outperforms PanoDreamer and Matrix3D by a clear margin, suggesting that the same geometry-aware parameterization transfers from projected wide-FoV views to native fisheye cameras.
For panoramic novel view synthesis, UniSHARP achieves the best PSNR, SSIM, and LPIPS on HM3D, OmniRooms, and the out-of-domain Replica dataset.
These results indicate that the proposed camera-unified design remains stable as the evaluation moves from wide-FoV and fisheye inputs to full $360^\circ$ panoramas.

\noindent\textbf{Pose-free performance.}
Table~\ref{tab:posefree_wildrgbd} evaluates the pose-free setting on WildRGB-D.
Ours uses the available camera parameters, while Ours (pose-free) estimates the camera model and rendering geometry from the predicted ray field.
The pose-free variant maintains competitive rendering quality without requiring camera calibration, demonstrating the practical value of ray-based camera recovery for unconstrained monocular inputs.

\subsection{Ablation Studies}

We conduct ablations on WildRGB-D and HM3D to analyze the main architectural and objective components of UniSHARP.

\begin{table*}[t]
\centering
\caption{Ablation study of model design components on WildRGB-D and HM3D. Each variant removes one component from the full model.}
\label{tab:ablation_model}
\resizebox{\linewidth}{!}{
\begin{tabular}{lcccccc}
\toprule
& \multicolumn{3}{c}{WildRGB-D~\citep{xia2024wildrgbd}} & \multicolumn{3}{c}{HM3D~\citep{ramakrishnan2021hm3d}} \\
\cmidrule(lr){2-4}\cmidrule(lr){5-7}
Variant & PSNR$\uparrow$ & SSIM$\uparrow$ & LPIPS$\downarrow$ & PSNR$\uparrow$ & SSIM$\uparrow$ & LPIPS$\downarrow$ \\
\midrule
Full model & \cellcolor{red!25}\textbf{21.56} & \cellcolor{red!25}\textbf{0.674} & \cellcolor{red!25}\textbf{0.143} & \cellcolor{red!25}\textbf{29.24} & \cellcolor{red!25}\textbf{0.895} & \cellcolor{red!25}\textbf{0.065} \\
w/o native resolution allocation & 21.21 & 0.664 & 0.155 & 28.72 & 0.872 & 0.071 \\
w/o second Gaussian layer & 20.63 & 0.642 & 0.168 & 28.29 & 0.877 & 0.083 \\
w/o panoramic distortion adaptation & 21.50 & 0.672 & 0.145 & 28.43 & 0.880 & 0.077 \\
w/ depth-RGB images input & 20.38 & 0.631 & 0.180 & 28.04 & 0.868 & 0.089 \\
\bottomrule
\end{tabular}}
\end{table*}

\noindent\textbf{Model design.}
Table~\ref{tab:ablation_model} summarizes the contribution of the main architectural components.
Replacing the proposed 2D semantic and 3D geometric features with direct depth-RGB inputs causes the largest degradation on both datasets, showing that the learned feature space provides stronger context than direct RGB-depth conditioning.
Removing the second Gaussian layer also hurts performance on both datasets, confirming the importance of an additional distance hypothesis for disocclusions and wide angular coverage.
Native resolution allocation remains important for preserving input detail, while panoramic distortion adaptation has a clearer effect on HM3D, where regularizing equirectangular distortion is more critical than in perspective images.

\section{Conclusion}
We presented UniSHARP, a universal-camera feedforward 3DGS framework for monocular novel view synthesis.
Starting from the observation that perspective-trained Gaussian regressors do not transfer reliably to heterogeneous camera systems, UniSHARP reformulates Gaussian prediction in a shared ray-distance space and composes Geometry Anchored Gaussians with Feature Conditioned Gaussian residuals.
This design preserves the efficiency of single-image Gaussian regression while supporting perspective, wide-FoV, fisheye, and panoramic inputs within one prediction model.
To evaluate this setting systematically, we further introduced a FoV stratified benchmark covering real and simulated scenes across narrow perspective to full panoramic cameras.
Experiments on this benchmark show that UniSHARP maintains strong performance on perspective datasets and substantially improves novel view synthesis across diverse camera systems.
We hope this work provides a practical foundation for monocular 3D Gaussian rendering in real-world imaging systems beyond the pinhole camera model.

\bibliographystyle{unsrtnat}
\bibliography{references}


\appendix

\section{Additional Experiments and Ablations}
\label{app:additional_experiments}

\subsection{Implementation Details}
\label{app:implementation_details}

All experiments are conducted on 8 H20 GPUs.
UniSHARP uses the feature-only architecture described in Sec.~3, with a UniK3D ViT-L backbone initialized from pretrained weights~\citep{piccinelli2025unik3d}.
For wide-FoV and fisheye rendering, we use the 3DGEER generic-camera Gaussian rasterizer~\citep{huang20263dgeer3dgaussianrendering}.
We optimize the model with Adam~\citep{kingma2014adam} for $10^6$ iterations using a $10^4$-iteration warmup followed by cosine learning-rate decay.
The learning rate decays from $1.0\times10^{-5}$ to $1.0\times10^{-6}$ for the depth head, and from $1.2\times10^{-4}$ to $1.6\times10^{-5}$ for the Gaussian decoder and prediction modules.
The loss weights are
$\lambda_c=1.0$, $\lambda_a=1.0$, $\lambda_p=1.0$, $\lambda_d=0.5$,
$\lambda_{\mathrm{tv}}=1.0$, $\lambda_g=0.5$, $\lambda_{\mathrm{gi}}=0.5$.

\subsection{Training Objective Ablation}
\label{app:loss_ablation}

\begin{table*}[h]
\centering
\caption{Ablation study of the main training losses on WildRGB-D and HM3D. Each variant removes one loss term from the full objective to measure its contribution to rendering quality.}
\label{tab:ablation_loss}
\resizebox{\linewidth}{!}{
\begin{tabular}{lcccccc}
\toprule
& \multicolumn{3}{c}{WildRGB-D~\citep{xia2024wildrgbd}} & \multicolumn{3}{c}{HM3D~\citep{ramakrishnan2021hm3d}} \\
\cmidrule(lr){2-4}\cmidrule(lr){5-7}
Variant & PSNR$\uparrow$ & SSIM$\uparrow$ & LPIPS$\downarrow$ & PSNR$\uparrow$ & SSIM$\uparrow$ & LPIPS$\downarrow$ \\
\midrule
Full objective & \textbf{21.56} & \textbf{0.674} & \textbf{0.143} & \textbf{29.24} & \textbf{0.895} & \textbf{0.065} \\
w/o perceptual appearance loss & 21.18 & 0.663 & 0.158 & 28.69 & 0.884 & 0.078 \\
w/o target rendered depth loss & 20.42 & 0.628 & 0.206 & 27.12 & 0.842 & 0.136 \\
w/o second-layer TV regularization & 21.30 & 0.667 & 0.151 & 28.56 & 0.879 & 0.084 \\
w/o floater suppression & 21.11 & 0.659 & 0.162 & 27.94 & 0.866 & 0.153 \\
\bottomrule
\end{tabular}}
\end{table*}

\noindent\textbf{Training objective.}
Table~\ref{tab:ablation_loss} analyzes the main loss terms.
The full objective achieves the best overall performance, reaching 21.56 PSNR and 0.143 LPIPS on WildRGB-D, and 29.24 PSNR and 0.065 LPIPS on HM3D.
Removing target rendered depth supervision causes the largest PSNR drop among the loss ablations, reducing PSNR to 20.42 on WildRGB-D and 27.12 on HM3D, while also increasing LPIPS to 0.206 and 0.136.
This shows that source-side depth supervision alone cannot constrain the Gaussian scene after view transformation; supervising the rendered target depth is critical for maintaining cross-view geometry and suppressing view-dependent distortions.
The perceptual appearance loss improves visual fidelity, while second-layer TV regularization and floater suppression stabilize the Gaussian field.
Floater suppression is particularly important for panoramic scenes, where removing it increases HM3D LPIPS from 0.065 to 0.153 due to unstable second-layer Gaussians near depth discontinuities.

\subsection{Fisheye Dataset Visualization}
\label{app:fisheye_visualization}

\begin{figure}[h]
    \centering
    \includegraphics[width=\linewidth]{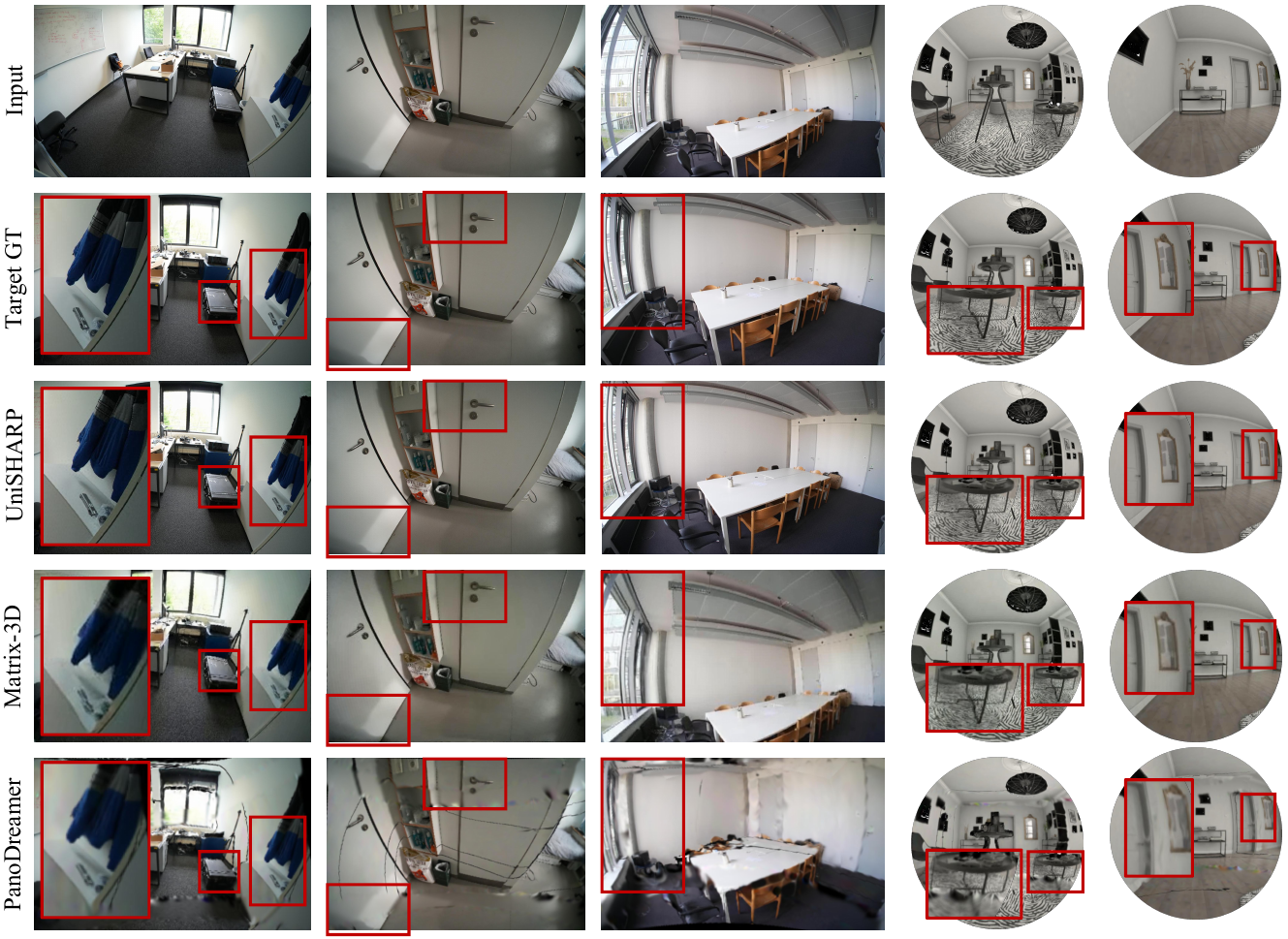}
    \caption{Visualization of the fisheye validation data used in our benchmark.
    The samples illustrate the strong radial distortion and wide angular coverage that distinguish native fisheye novel view synthesis from standard perspective evaluation.
    }
    \label{fig:fish_dataset}
\end{figure}

\subsection{Panoramic Inference via Cubemap Decomposition}
\label{app:cubemap_sharp}

As discussed in Sec.~1, SHARP~\citep{mescheder2026sharp} maps every pixel in normalized image space under a pinhole camera assumption and therefore cannot directly ingest equirectangular panoramas or other non-pinhole inputs.
A common workaround is to decompose the panorama into six cubemap faces, run SHARP independently on each face, and then fuse the resulting Gaussian predictions before rendering back to the panoramic domain.
However, this pipeline inherits the limitations noted in the introduction: each face is processed in isolation under a different local pinhole approximation, so the predicted Gaussian fields are not globally consistent across face boundaries.
When the per-face renderings are stitched into a full panorama, the inconsistency manifests as prominent seams and view-dependent discontinuities, especially near geometric edges and depth boundaries.

\begin{figure}[h]
    \centering
    \includegraphics[width=\linewidth]{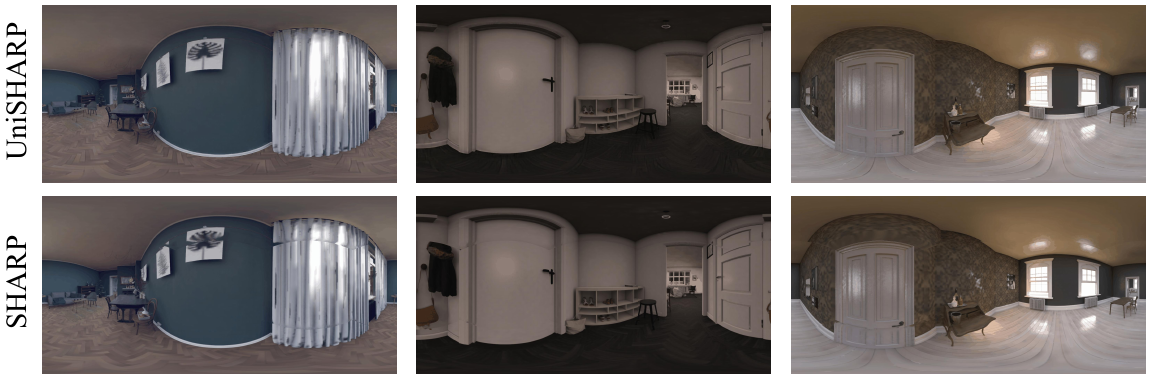}
    \caption{Comparison of panoramic novel view synthesis with a cubemap-based SHARP baseline and UniSHARP.
    For SHARP, we split the source panorama into cubemap faces, run feedforward inference on each face separately, and stitch the rendered target views back into equirectangular format.
    The stitched result exhibits large cubemap seams caused by inconsistent cross-face geometry.
    In contrast, UniSHARP operates directly on the panoramic input in unified ray-distance space and produces a seamless target rendering without cubemap decomposition or post-stitching.
    }
    \label{fig:cubesharp}
\end{figure}

Figure~\ref{fig:cubesharp} visualizes this failure mode on a representative panoramic sample.
The cubemap-based SHARP baseline produces clearly visible stitching artifacts at face junctions, whereas UniSHARP renders a coherent panoramic target view without such seams.
This comparison supports our design choice to avoid pinhole-specific re-projection heuristics and instead predict Gaussians in a camera-unified representation that remains valid across the full $360^\circ$ field of view.

\subsection{Inference Time Comparison}
\label{app:runtime}

\begin{table}[h]
\centering
\caption{Inference time comparison for single-image novel view synthesis. Runtime is measured under the same evaluation setting; relative runtime is reported with respect to UniSHARP.}
\label{tab:runtime}
\begin{tabular}{lcc}
\toprule
Method & Runtime $\downarrow$ & Relative runtime $\downarrow$ \\
\midrule
UniSHARP & \textbf{3.1s} & -- \\
PanoDreamer~\citep{paliwal2025panodreamer} & 8.6s & 2.8$\times$ \\
Matrix3D~\citep{lu2025matrix3d} & 38.8s & 12.5$\times$ \\
\bottomrule
\end{tabular}
\end{table}

\noindent\textbf{Inference efficiency.}
Table~\ref{tab:runtime} compares the inference time of UniSHARP with panoramic baselines.
UniSHARP completes inference in 3.1 seconds, while PanoDreamer and Matrix3D require 8.6 seconds and 38.8 seconds, respectively.
PanoDreamer and Matrix3D are therefore 2.8$\times$ and 12.5$\times$ slower than UniSHARP.
The speed advantage comes from the model design: UniSHARP predicts the complete Gaussian representation with a single feedforward pass and directly renders it, avoiding per-scene optimization, diffusion sampling, or iterative video generation at inference time.

\section{Limitations}
\label{app:limitations}

UniSHARP is a feedforward Gaussian prediction model rather than a generative scene completion method.
With mixed-camera training, the model can acquire a degree of extrapolation ability and can handle moderate disocclusions around the input view.
However, when target views expose large regions that are completely outside the source image, the model has limited evidence for hallucinating unseen content.
In such cases, holes or weakly supported structures may appear near outer boundary regions.
Improving long-range extrapolation while preserving the efficiency and geometric consistency of feedforward Gaussian prediction remains an interesting direction for future work.

\section{Societal Impact}
\label{app:societal_impact}

Universal-camera monocular novel view synthesis can broaden the use of 3D perception and rendering systems beyond standard perspective imagery.
By supporting perspective, wide-FoV, fisheye, and panoramic inputs in a unified model, UniSHARP may benefit applications such as embodied AI, robotics, AR/VR content creation, immersive telepresence, and spatial documentation.
The ability to infer renderable 3D structure from a single image can reduce capture requirements and make 3D content generation more accessible for devices with diverse camera systems.
The proposed benchmark and dataset can also support more systematic evaluation of camera-general view synthesis methods, encouraging future research on robust and geometry-aware spatial intelligence.

\section{Benchmark Details}
\label{app:benchmark}

This section supplements the benchmark description in Sec.~\ref{sec:benchmark}. The main paper introduces the benchmark composition and reports results across perspective, wide-FoV, fisheye, and panoramic camera groups. Here we provide additional details on data splits, pair construction, camera metadata, metric computation, baseline adaptation, and evaluation protocol, so that the benchmark can be reproduced and reused by future work.

\subsection{Dataset Splits and Scene Selection}
\label{app:benchmark_splits}

For all datasets, validation samples are constructed in a single-source multi-target format. Each sample contains one source image, one or more target views, the corresponding camera parameters when available, and metadata describing the projection type and effective FoV. Target-view RGB images are used only for evaluation and are never provided to the model during inference.

For existing datasets, we follow the official validation or test splits whenever they are available. When a dataset does not provide a standard split for monocular novel view synthesis, we construct a held-out validation split at the sequence or scene level to avoid source-target leakage across training and evaluation. The evaluation samples are fixed before testing, which makes the benchmark deterministic and avoids dependence on dataloader randomness.

\paragraph{Perspective datasets.}
For RealEstate10K, DL3DV, Tanks and Temples, and WildRGB-D, we use the perspective camera metadata provided by the original datasets. RealEstate10K follows its official test split, while the remaining datasets use held-out scene or sequence subsets constructed for monocular novel view synthesis. Tanks and Temples is used only as a held-out perspective evaluation set to measure out-of-domain generalization.

\paragraph{Wide-FoV dataset.}
OmniRooms-Wide is derived from OmniRooms by projecting equirectangular panoramas into wide-FoV views. We use a $130^\circ$ equidistant projection at $1024\times1024$ resolution. Source and target views within the same local group share the same camera orientation, so the benchmark isolates translation-induced view synthesis rather than mixing translation and rotation. The metadata records the projection type, FoV, valid image radius, and camera-to-world transform for each rendered view.

\paragraph{Fisheye dataset.}
For ScanNet++ Fisheye, we preserve the native fisheye camera model and use the provided camera calibration when available. Frames without valid calibration or depth are skipped before pair construction. 

\paragraph{Panoramic datasets.}
For HM3D, Replica, and OmniRooms, images are evaluated in equirectangular projection. Replica is used as an out-of-domain panoramic validation set, while HM3D and OmniRooms evaluate panoramic rendering quality under real-scanned and simulated indoor scenes, respectively. The OmniRooms panoramic split is generated from fixed simulated trajectories, and all source-target groups are defined before evaluation. The complete scene and frame identifiers are provided with the released benchmark metadata.

\subsection{OmniRooms Construction}
\label{app:omnirooms}

OmniRooms is a simulated indoor equirectangular panorama dataset built to provide dense local camera trajectories for monocular panoramic novel view synthesis. For each valid anchor location, we render one central source camera and multiple nearby target cameras within a local neighborhood. All cameras associated with the same anchor share the same orientation, so the benchmark isolates translation-induced view synthesis rather than mixing translation and rotation. Each panorama is rendered at $1024 \times 2048$ resolution.

\paragraph{Anchor sampling.}
Anchor locations are sampled on a $0.5$m voxel grid in navigable indoor regions before local camera expansion. We retain only anchor centers whose height coordinate satisfies $60 \leq Z \leq 180$ cm, which removes floor-level, ceiling-level, and otherwise implausible camera centers. Collision, navigability, and degenerate-rendering checks are applied during simulation and data filtering. A rendered sample is discarded if its RGB image, depth map, camera metadata, or source-target visibility overlap does not satisfy the benchmark requirements.

\paragraph{Target sampling.}
For each retained anchor, target cameras are sampled in a local axis-aligned cube of edge length $30$ cm around the source camera. Each anchor produces 30 camera positions: the original center and 29 randomly perturbed target centers. This local sampling range evaluates nearby-view synthesis while still introducing meaningful parallax and disocclusions.

\paragraph{Rendering settings.}
Each OmniRooms sample contains an RGB panorama, aligned depth, and camera metadata. Panoramic RGB images are rendered at $1024 \times 2048$ resolution. Depth and surface geometry are used for supervision, visibility filtering, and analysis, but are not provided as model input at test time. RGB values are decoded and normalized to $[0,1]$ before evaluation.

\subsection{Source-Target Pair Filtering}
\label{app:pair_filtering}

The benchmark focuses on local monocular novel view synthesis. We therefore filter source-target pairs to avoid evaluating unconstrained long-range hallucination. A pair is retained only if it satisfies three constraints: source-target overlap at least $60\%$, camera-center distance smaller than $0.5$m, and image-index gap at most 10. These constraints focus the evaluation on geometry and disocclusion reasoning under meaningful local motion.

\paragraph{Camera distance.}
Camera-center distance is computed as the Euclidean distance between the source and target camera centers in the dataset coordinate system:
\begin{equation}
    d(s,t) = \| \mathbf{c}_s - \mathbf{c}_t \|_2 .
\end{equation}
The pair is valid if $d(s,t) < 0.5$m.

\paragraph{Frame-index gap.}
For sequence-based datasets, the index gap is computed using the original frame order. For real-world video or image-sequence datasets, frames follow the temporal or acquisition order provided by the original data. For simulated panoramic data, frames are ordered according to the generated local camera groups. For datasets without native video order, source and target indices are fixed by the benchmark metadata.

\paragraph{Source-target overlap.}
Overlap is measured as the fraction of target-view pixels whose corresponding visible 3D points are also visible in the source view. When ground-truth depth is available, we compute overlap by back-projecting target pixels into 3D and re-projecting them into the source camera. A projected point is counted as overlapping if it lies inside the source image domain and passes a visibility check. For panoramic data, overlap is computed with circular horizontal wrap-around. For fisheye data, the native fisheye valid mask is applied before counting pixels. We use dataset-provided depth, camera poses, or meshes for this filtering.

A generic overlap computation can be written as:
\begin{equation}
    \mathrm{Overlap}(s,t) =
    \frac{1}{|\Omega_t|}
    \sum_{\mathbf{p}\in\Omega_t}
    \mathbf{1}
    \left[
    \Pi_s
    \left(
    \Pi_t^{-1}(\mathbf{p}, D_t(\mathbf{p}))
    \right)
    \in \Omega_s
    \right],
\end{equation}
where $\Omega_s$ and $\Omega_t$ denote source and target image domains, $D_t$ is the target depth map, and $\Pi_s$, $\Pi_t$ are the corresponding camera projection functions.

\subsection{Camera Metadata and Projection Models}
\label{app:camera_metadata}

Each benchmark sample is associated with camera metadata. We store camera parameters in a unified format while preserving the native projection model of each dataset. Extrinsics are represented as camera-to-world transforms after converting dataset-specific coordinate conventions to the common training convention used by UniSHARP.

\paragraph{Perspective cameras.}
Perspective samples use pinhole intrinsics:
\begin{equation}
    \mathbf{K} =
    \begin{bmatrix}
    f_x & 0 & c_x \\
    0 & f_y & c_y \\
    0 & 0 & 1
    \end{bmatrix}.
\end{equation}

\paragraph{Wide-FoV cameras.}
OmniRooms-Wide uses an equidistant wide-FoV projection:
\begin{equation}
    r = f \theta,
\end{equation}
where $\theta$ is the angle between the optical axis and the viewing ray, and $r$ is the radial distance from the image center. For the $130^\circ$ rendered views, the metadata records the FoV, image size, valid radius, and per-frame camera transform.

\paragraph{Fisheye cameras.}
For fisheye datasets, we preserve the native fisheye calibration provided by the dataset whenever available. The benchmark stores the corresponding fisheye projection parameters and converts them to a renderer-compatible representation during evaluation. For simulated fisheye views, we use an equidistant fisheye projection with a $130^\circ$ FoV, $1024\times1024$ resolution, and a valid radius of 512 pixels.

\paragraph{Panoramic cameras.}
Panoramic samples use equirectangular projection. For a pixel coordinate $(u,v)$ in an image of width $W$ and height $H$, longitude and latitude are computed as:
\begin{equation}
    \phi = 2\pi \left(\frac{u + 0.5}{W}\right) - \pi,
    \quad
    \theta = \frac{\pi}{2} - \pi \left(\frac{v + 0.5}{H}\right).
\end{equation}
The corresponding unit ray is:
\begin{equation}
    \mathbf{r} =
    \begin{bmatrix}
    \cos\theta \sin\phi \\
    \sin\theta \\
    \cos\theta \cos\phi
    \end{bmatrix}.
\end{equation}

\subsection{Evaluation Protocol}
\label{app:evaluation_protocol}

The benchmark follows a single-source multi-target protocol. For each sequence, the first frame in a predefined evaluation group is used as the source view, and target views are selected from subsequent valid frames whose frame-index gap is at most 10. The model receives only the source image and, in the calibrated setting, the source and target camera parameters. No target-view RGB information is used during inference.

For each valid target view, the method renders an image in the target camera projection. The rendered image is compared with the ground-truth target image using PSNR, SSIM, and LPIPS. Metrics are first averaged over all valid target views of a source sequence, then averaged over sequences within each dataset. We report dataset-level scores in the main paper.

\paragraph{Resolution.}
All methods are evaluated at the ground-truth target resolution used by the corresponding validation sample, unless a fixed validation resizing multiple is specified. In that case, both predictions and targets are resized consistently before metric computation. OmniRooms equirectangular panoramas use $1024\times2048$ resolution. Perspective datasets are evaluated at the image resolution produced by the corresponding dataset loader after the same deterministic resizing rule.

\paragraph{Valid masks.}
If a dataset contains invalid pixels, missing depth, black borders, or undefined fisheye regions, metrics are computed only on valid pixels. Fisheye views use the valid fisheye mask, and wide-FoV equidistant views use the valid rendered image domain recorded by the camera metadata. For no-extrapolation rendering modes, border-connected black invalid regions are excluded from metric computation.

\paragraph{Panoramic boundary handling.}
For equirectangular images, horizontal coordinates are circular. During rendering and any geometric filtering, longitude wrap-around is handled circularly. For image-quality metrics, predictions and ground-truth images are compared in the same equirectangular coordinate system.

\subsection{Metric Implementation}
\label{app:metric_implementation}

We use PSNR, SSIM, and LPIPS as the benchmark metrics. PSNR measures pixel-level reconstruction fidelity, SSIM measures structural similarity, and LPIPS measures perceptual similarity.

\paragraph{PSNR.}
PSNR is computed from the mean squared error over valid pixels:
\begin{equation}
    \mathrm{PSNR} = -10 \log_{10}(\mathrm{MSE}),
\end{equation}
assuming RGB values are normalized to $[0,1]$.

\paragraph{SSIM.}
SSIM is computed on RGB images using a Gaussian window of size 11 and standard deviation 1.5. We use constants $C_1=0.01^2$ and $C_2=0.03^2$ on normalized RGB values. Images are not converted to luminance; the RGB-channel SSIM values are averaged.

\paragraph{LPIPS.}
LPIPS is computed using the official \texttt{lpips} implementation with the AlexNet backbone, i.e., \texttt{LPIPS(net="alex")}. All benchmark numbers reported in the paper use this LPIPS-Alex setting.

\paragraph{Aggregation.}
For each dataset, metrics are averaged across target views and then across source sequences. Camera-group averages are computed as unweighted averages over dataset-level scores in the same camera group, so large validation sets do not dominate the group score.

\subsection{Baseline Evaluation Details}
\label{app:baseline_details}

We evaluate all baselines on the same source-target pairs as UniSHARP. For each baseline, we use official checkpoints and inference code when available. If a method supports only a subset of camera models, we use the closest compatible input representation and keep the target-view evaluation protocol unchanged.

\paragraph{Perspective baselines.}
SHARP, Flash3D, LVSM, and TMPI are evaluated on perspective datasets. These methods are designed primarily for perspective monocular view synthesis, so we use the original perspective camera parameters and render target views under the corresponding pinhole cameras. Inputs are resized using the official preprocessing of each method, and outputs are resized back to the target resolution before metric computation.

\paragraph{Wide-FoV, fisheye, and panoramic baselines.}
PanoDreamer and Matrix3D are evaluated on non-perspective datasets because they support broader view synthesis or panoramic generation settings. For each method, we convert the benchmark input into the representation expected by the official implementation and then render or sample the corresponding target views. Camera trajectories are always taken from the benchmark metadata. We use the official default number of diffusion steps, optimization iterations, and post-processing settings unless the method cannot produce the required projection directly, in which case the output is first rendered in the method's native projection and then reprojected to the benchmark target camera.

\paragraph{Runtime measurement.}
For runtime comparisons, all methods are evaluated under the same hardware setting and using the same number of rendered target views. Timing excludes dataset loading and image decoding, includes model forward and target-view rendering, and is measured after warm-up runs. Batch size, image resolution, GPU type, and the number of warm-up and measured iterations are fixed across methods.

\subsection{Pose-Free Evaluation Details}
\label{app:pose_free_eval}

The pose-free setting evaluates whether a method can operate without manually provided camera intrinsics. In this setting, the model receives only the source RGB image. UniSHARP predicts a ray field, infers the camera model, and recovers rendering geometry before synthesizing target views.

We evaluate pose-free rendering on WildRGB-D. The calibrated setting uses the available camera parameters, while the pose-free setting replaces the source camera calibration with the recovered camera geometry. Target camera parameters are still provided to define the evaluation views and to make the rendered images comparable with ground-truth target frames. Thus the pose-free setting removes source-camera calibration from the model input, but does not change the target-view definitions used for metric computation.

\subsection{Quality Control}
\label{app:quality_control}

We apply the filtering criteria introduced in Sec.~4.1 and further remove samples with missing source or target images, invalid camera metadata, or missing/invalid depth when depth is required for overlap computation. Fisheye samples are evaluated only inside the valid fisheye mask. In no-extrapolation modes, connected black border regions introduced by rendering are excluded from the valid metric mask.

For simulated OmniRooms data, we additionally remove samples with invalid depth buffers or severe clipping. The camera-position expansion step keeps only centers with $60 \leq Z \leq 180$ cm. Because each generated group starts from a retained center and clips the generated $Z$ coordinate to the same valid range, all released local camera positions remain within the configured height interval.

\subsection{Licenses, Ethics, and Privacy}
\label{app:benchmark_ethics}

The benchmark combines existing public datasets with the newly constructed OmniRooms and OmniRooms-Wide data. For existing datasets, users should follow the licenses and usage terms of the original datasets. OmniRooms and OmniRooms-Wide are released under the CC BY-NC 4.0 license for research and non-commercial use. The released metadata will not include private user information.



\end{document}